\documentclass[runningheads,a4paper]{llncs}[2015/06/24]

\usepackage{cmap}

\iffalse % use default-font
\usepackage[%
rm={oldstyle=false,proportional=true},%
sf={oldstyle=false,proportional=true},%
tt={oldstyle=false,proportional=true,variable=true},%
qt=false%
]{cfr-lm}
\else
\usepackage{newtxtext}
\usepackage{newtxmath}
\usepackage[zerostyle=b,scaled=.9]{newtxtt}
\fi

\usepackage[math]{blindtext}
\usepackage{mwe}

\usepackage[T1]{fontenc}
\usepackage[utf8]{inputenc} %support umlauts in the input

\usepackage{graphicx}

\usepackage[ngerman,english]{babel}
\addto\extrasenglish{\languageshorthands{ngerman}\useshorthands{"}}

\usepackage{upquote}

\usepackage{booktabs}

\usepackage{paralist}

\usepackage{csquotes}

\RequirePackage{iftex}
\ifPDFTeX
\RequirePackage[%
final,%
expansion=alltext,%
protrusion=alltext-nott]{microtype}%
\else
\RequirePackage[%
final,%
protrusion=alltext-nott]{microtype}%
\fi%
\DisableLigatures{encoding = T1, family = tt* }

\usepackage{url}
\makeatletter
\g@addto@macro{\UrlBreaks}{\UrlOrds}
\makeatother
\usepackage[usenames,dvipsnames,svgnames,table]{xcolor}

\usepackage{listings}
\renewcommand{\lstlistingname}{List.}
\usepackage{chngcntr}
\AtBeginDocument{\counterwithout{lstlisting}{section}}

\AtBeginEnvironment{listing}{\setcounter{listing}{\value{lstlisting}}}
\AtEndEnvironment{listing}{\stepcounter{lstlisting}}

\usepackage{pdfcomment}
\usepackage[%
square,        % for square brackets
comma,         % use commas as separators
numbers,       % for numerical citations;
sort&compress, % as sort but in addition multiple numerical citations
]{natbib}
\SetExpansion
[ context = sloppy,
stretch = 30,
shrink = 60,
step = 5 ]
{ encoding = {OT1,T1,TS1} }
{ }

\usepackage[capitalise,nameinlink]{cleveref}
\crefname{section}{Sect.}{Sect.}
\Crefname{section}{Section}{Sections}
\crefname{listing}{\lstlistingname}{\lstlistingname}
\Crefname{listing}{Listing}{Listings}

\usepackage{ltxcmds}
\makeatletter
\newcommand{\IfPackageLoaded}[2]{\ltx@ifpackageloaded{#1}{#2}{}}
\makeatother

\IfPackageLoaded{minted}{
	\setminted{numbersep=5pt, xleftmargin=12pt}
	
	\usemintedstyle{friendly}
	
	\usepackage[labelfont=bf,font=small,skip=4pt]{caption}
	\SetupFloatingEnvironment{listing}{name=List.,within=none}
}{
}
\ifcsmacro{listing}{}{
	\newenvironment{listing}[1][htbp!]{\begin{figure}[#1]}{\end{figure}}
	\newcounter{listing}
}
\ifcsmacro{minted}{}{%
	
}

\usepackage{xspace}
\DeclareFontFamily{U}{MnSymbolC}{}
\DeclareSymbolFont{MnSyC}{U}{MnSymbolC}{m}{n}
\DeclareFontShape{U}{MnSymbolC}{m}{n}{
	<-6>    MnSymbolC5
	<6-7>   MnSymbolC6
	<7-8>   MnSymbolC7
	<8-9>   MnSymbolC8
	<9-10>  MnSymbolC9
	<10-12> MnSymbolC10
	<12->   MnSymbolC12%
}{}
\DeclareMathSymbol{\powerset}{\mathord}{MnSyC}{180}

\usepackage[caption=false,font=normalsize,labelfont=sf,textfont=sf]{subfig}
\usepackage{capt-of}

\newcommand{\etal}{\emph{et al.}}

\newcommand{\figref}[1]{Fig.~\ref{#1}}
\newcommand{\secref}[1]{Section~\ref{#1}}

\newcommand{\formulainstrumentposition}{\vec{x}_{instr}}

\newcommand{\formulaorientangle}{\alpha}

\usepackage{siunitx}

\begin{document}
	
	\title{How Bad is Good enough%\thanks{Grants or other notes
	}
	\subtitle{Noisy annotations for instrument pose estimation}
	
	
	\author{%
		David Kügler         \and
		%        Jannik Sehring       \and %etc.
		%        Tim Bergmann         \and
		        Anirban Mukhopadhyay
	}
	
	
	\institute{Interactive Graphics Systems Group, Technische Universität Darmstadt}
	
	%% Multiple insitutes - ALTERNATIVE to the above
	% \author{%
	%     Firstname Lastname\inst{1} \and
	%     Firstname Lastname\inst{2}
	% }
	%
	%If there are too many authors, use \authorrunning
	%  \authorrunning{First Author et al.}
	%
	%  \institute{
	%      Insitute 1\\
	%      \email{...}\and
	%      Insitute 2\\
	%      \email{...}
	%}
	
	\maketitle
	
	\begin{abstract}
		Though analysis of Medical Images by Deep Learning achieves unprecedented results across various applications, 
		the effect of \emph{noisy training annotations} is rarely studied in a systematic manner. 
		In Medical Image Analysis, most reports addressing this question concentrate on studying segmentation performance of deep learning classifiers.
		The absence of continuous ground truth annotations in these studies limits the value of conclusions for applications, where regression is the primary method of choice. 
		In the application of surgical instrument pose estimation, where precision has a direct clinical impact on patient outcome, studying the effect of \emph{noisy annotations} on deep learning pose estimation techniques is of supreme importance. 
		Real x-ray images are inadequate for this evaluation due to the unavailability of ground truth annotations. 
		We circumvent this problem by generating synthetic radiographs, where the ground truth pose is known and therefore the pose estimation error made by the medical-expert can be estimated from experiments. 
		Furthermore, this study shows the property of deep neural networks to learn dominant signals from noisy annotations with sufficient data in a regression setting.  
	\end{abstract}
	
	
%input{text/example-tex}
	
% image sizes:
% \textwidth ~ 122 mm
% 1.5* \textwidth ~ 183 mm
% half col: ~ 89 mm
	
\section{Introduction}
\label{sec:intro}
\begin{figure}[t]
	\centering
	\includegraphics[width=\textwidth]{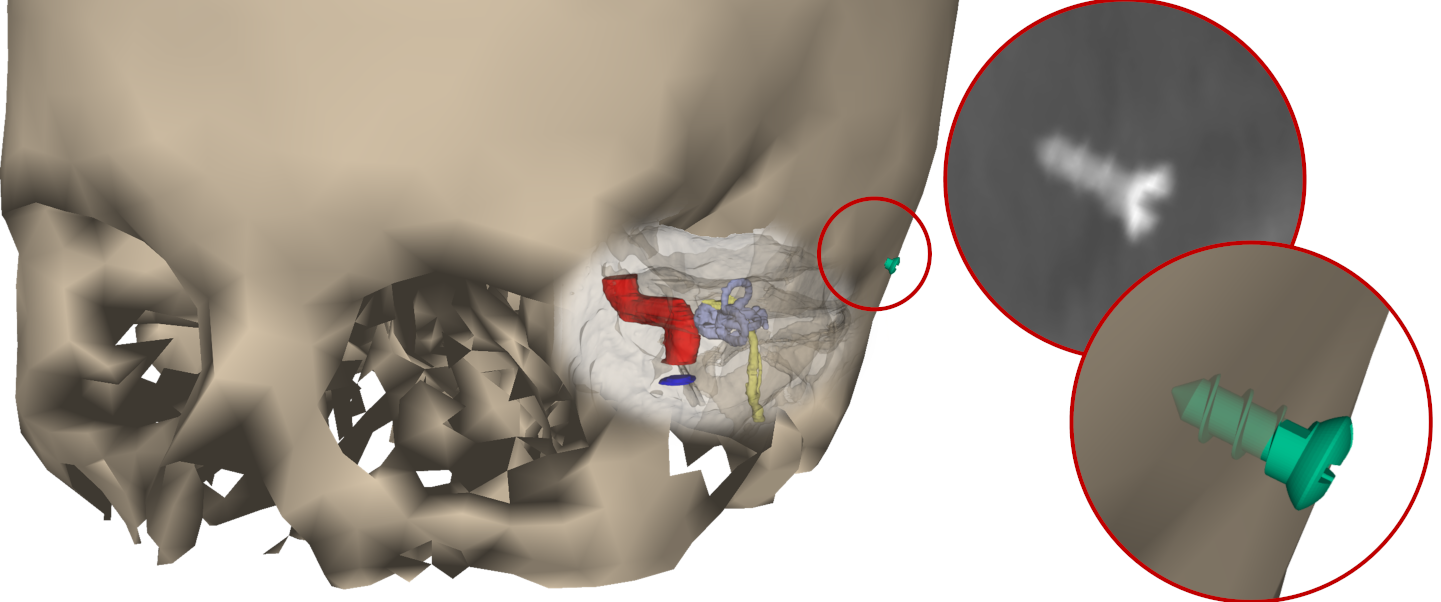}
	\caption{\label{fig:overview}The window into temporal bone illustrates several risk structures. A safety margin to these structures including nerves (yellow), arteries (red) and cochlea and labyrinth (blue) has to be ensured in otobasis surgery. 
		We use a screw (green) displayed at scale as a prototypical miniaturized surgical instrument.
		Generated x-ray images show screw and anatomy for the pose estimation task.}
\end{figure}

The advantages of Minimally Invasive Surgery over conventional surgery are reduced trauma and increased patient security \cite{Koffron.2007}.
To achieve this, the precise navigation of the instrument along a path ensures a safety margin to risk structures. 
Especially for the scenario of otobasis surgery (see \figref{fig:overview}) investigated in this paper,
the close proximity of critical risk structures, such as cochlea and nerves, calls for high-precision navigation \cite{Schipper.2004}.
%Automatic and computer-aided strategies in combination with robotic instruments increasingly improve the navigation quality.
Conventional tracking systems require line-of-sight (optical tracking) or provide low accuracy (electromagnetic tracking).
Upcoming techniques based on the combination of radiographs (using a c-arm) and Deep Learning (such as i3PosNet \cite{Kugler.20180226}) are independent of these limitations.
Studies for manual instrument poses estimation lack either a systematic analysis, the comparison to ground truth values, users not being medical experts (surgeons) and/or the compatibility with the presented application. 
Therefore the usability of manual annotations as ground truth is an open question.

%By increases in instrument pose accuracy with radiographs from ,
%these advantages can further improve the climical outcome, 
%The measurement accuracy of the instrument is a major challenge and potent lever to improve the placement accuracy by automatic control and therefore the clinical outcome.
%Since conventional tracking systems are limited by the required line-of-sight (optical tracking) or the provided accuracy (electromagnetic tracking), 
%we are interested in alternative methods based on images such as radiographs.
%These current c-arms (such as Ziehm Vision RFD) provide high-resolution x-ray images largely independent of line-of-sight.

In analogy to most Medical Image Analysis (MIA) problems, the introduction of Deep Learning (DL) for instrument pose estimation has shown significant performance gains compared to previous registration based methods \cite{Kugler.20180226,Kugler.2018,Miao.2016}, while trained on ground truth annotations. 
%Despite only covering a small percentage of possible images, 
%they generalize well to unseen cases (generalization error).
%Introducing and inaccurate annotations impose additional strain on the generalization capabilities of Deep Neural Networks.
Because of the additional dependence of these techniques on \emph{training annotations}, the annotation quality might influence the performance.
In fact, previous work on annotation quality focused on the effects for classification \cite{Frenay.2014,Rolnick.20180226} and segmentation \cite{Albarqouni.2016,Jungo.20180607} -- albeit under no ground truth knowledge.
However, such a relationship remains unexplored for instrument pose estimation (and regression) in MIA.
%Achievable reference accuracies are significantly less accurate than pose estimation by DL (i3PosNet \cite{Kugler.20180226}: \SI{0.031}{\milli\meter})
%independent of whether annotations are gained manually by experts (approx. \SI{0.5}{\milli\meter}, see Chapter 4) 
%or experimentally by reference measurements (coordinate measurement machine \cite{ZEISSCenterMax.2017}: $MPE_{E0}$ approx. \SI{0.2}{\milli\meter} at length \SI{60}{\milli\meter}).

%where overfitting to these samples might negatively influence above appreciated generalization property.
%%%

This paper connects a MIA problem from Minimally Invasive Surgery (instrument pose estimation) to an ongoing discourse in the Learning Community (Generalization Theory) by studying the effect \emph{noisy annotations} have on the generalization error. 
We examine a) the quality of pose estimations generated by medical experts in order to investigate b) the \emph{effect of the noise level} on the prediction accuracy and c) how the \emph{number of images} with noisy annotations impacts performance in comparison with medical experts, which has never been investigated before.

For our analysis, we generate Digitally Rendered Radiographs (DRRs) from a CAD model and CTs of human cadavers. 
The CAD model implements a prototypical surgical instrument, which is a screw in this evaluation.
As a result, unlike in the real world scenario, we can compare clinical expert's performance with i3PosNet's where the Ground Truth is precisely known.
a) We design a Graphical User Interface and evaluate pose annotation errors of six medical experts. 
Our analysis of the pose estimation performance of medical experts reveals the limited capability of humans to estimate the pose of instruments.
The orientation is especially difficult to estimate.
This analysis motivated the investigation of \emph{noisy annotations}.
b) We evaluate the performance of i3PosNet on noisy annotations.
The results indicate a linear impact of the annotation noise level on the generalization error.
c) Finally, we analyze i3PosNet's performance based on annotation noise at a human level for different dataset sizes and compare to the quality of human annotations.
Given a sufficiently large dataset, the performance even exceeds the noise level of the annotations.

%In this paper, we present two contributions:
%\begin{itemize}
%	\item an analysis of the pose estimation accuracy of humans (medical experts) and
%	\item a comparison of the Deep Learning pose estimation framework i3PosNet with human performance.
%	%	\item idea for a cooperative strategy
%\end{itemize}

\section{Related Work}
\label{sec:relatedwork}

Frenay \etal \cite{Frenay.2014} review the robustness of classification to noisy labels (annotations) and propose a categorization of ``label noise-robust'', ``label noise cleansing'', and ``label noise-tolerant'' methods.
%\needswork{voting scheme, expectation maximization}
%\needswork{uncertainty estimation, Yarin}
Rolnick \etal \cite{Rolnick.20180226} show neural networks can be robust for certain classification problems even to high noise.
For segmentation, Albarqouni \etal \cite{Albarqouni.2016} proposed a voting scheme as part of their neural network to address noisy annotations created by non-experts in the cloud.
Jungo \etal \cite{Jungo.20180607} analyze the effect of uncertainty estimation with respect to inter-observer variability.
How well Deep Neural Networks handle noisy annotations for regression is not addressed for MIA. %in the context of Computer-Aided Interventions (CAI).

While Registration still is the most frequently used method for automatic pose estimation of instruments \cite{Uneri.2015,Kugler.2018,Hatt.2016,Esfandiari.2016,Jain.2005}, 
several Learning-based methods have recently been introduced.
% \cite{Vandini.2017b} uses local features (SEGlets)
Miao \etal \cite{Miao.2016} introduced Learning into the registration pipeline using multiple specialized convolutional neural networks to replace the common registration components metric and optimizer, but still generate DRRs on-the-fly.
Bui \etal \cite{Bui.2017} shows that Deep Neural Networks predict projection parameters from x-rays outperforming other regression approaches.
While Bui \etal{} only calculate DRRs before and during training, they do not consider an anatomy surrounding the object.
Finally, Kügler \etal \cite{Kugler.20180226} introduce geometric considerations with a modular Deep Learning Framework (i3PosNet), that is anatomy independent and works for multiple instruments and reframes the pose estimation problem as a regression problem. 

Target registration error is an often used metric to evaluate pose estimation experiments independent of ground truth poses in images, but it requires custom phantoms often incompatible with surrounding anatomy, multiple configurations and still subject to the phantoms calibration accuracy.
But, unlike registration, regression-based methods require individual ground truth poses bringing unaccounted errors into the learning process for internal models.
The generation of ground truth poses has been addressed by manual registration \cite{Kaiser.2014}, calibrated setups \cite{Bui.2017}, comparison with reference methods \cite{Miao.2016} and artificially generating images \cite{Bui.2017,Miao.2016,Kugler.20180226}.

\section{Materials}
\label{sec:materials}
For our analysis, we rely on Digitally Rendered Radiographs (DRRs), because only for those real ground truth poses are available, which is ideal for systematically studying the regression performance of CNNs in our application environment.

% 0.25 pages
\textbf{Radiographs: }\label{sec:radiographs}
We generate multiple sets of radiographs for different use-cases combining real anatomies and the instrument mesh (number of independent DRRs): evaluation by medical experts (200), training (10\,k), validation (2\,k) and experiments (1\,k).
From three scanned anatomies (spacing from \SI{0.37}{\milli\meter} to \SI{0.44}{\milli\meter} and \SI{0.6}{\milli\meter} between slices), we train and validate i3PosNet in a leave-one-anatomy-out design: While two anatomies are used for training and validation of i3PosNet, the third is reserved for experiments yielding generalization to the anatomy.
Twenty clinically plausible poses per anatomy provide the references, in whose vicinity we draw sample poses for our screw to be placed.
These references are located on both sides around the ear on the skull.

The geometry parameters for the DRR generation follow our in-house Ziehm Vision RFD c-arm setup to ensure realistic instrument sizes and resolutions in the image (see \figref{fig:overview}).
Ground truth poses of the screw are calculated by projecting the origin (head of the screw, which is typically used for pivot calibration) onto the detector according to this projection geometry.
Of this pose, we consider the position on the image (pixel-coordinates, $\formulainstrumentposition$) as well as the angle between the projected instrument's main axis and the images x-axis (forward angle, $\formulaorientangle$) for the evaluation.

\section{Methods}
\label{sec:methods}

We determine which errors we should expect of x-ray images manually annotated by medical experts and derive the associated distribution.
By simulating this noise we obtain larger data sets with consistent statistical properties.
Applying i3PosNet \cite{Kugler.20180226} to the dataset, we analyze the effect of noisy annotations on i3PosNet -- a Deep Learning pose estimation framework.

%\begin{figure}[b]
%	% Use the relevant command to insert your figure file.
%	% For example, with the graphicx package use
%	\centering
%	\includegraphics[width=0.75\textwidth]{ScreenTool5-1}
%	% figure caption is below the figure
%	\caption{Tool for expert interaction and screw pose identification}
%	\label{fig:interactivetool}       % Give a unique label
%\end{figure}
\textbf{Interactive Tool for pose estimation: }
\label{sec:interactivetool}
We created a tool to manually identify the pose of surgical instruments. 
Poses have 5 degrees of freedom: 3-dimensional world coordinates as well as pitch and yaw. While roll can be adjusted in the interface, its only use is for visual representation.
Two projections of the same scenario share a split screen allowing the user to overcome the difficulty of deriving the depth from a single image by simultaneously viewing the annotation from two projection directions.
In the viewports, the DRRs are superimposed with the outline of the screw in the annotated position, which can be manipulated with the mouse. 

We rely on the perfect knowledge of the projection parameters to properly construct the scene, 
which is available since we use simulated x-rays.% (\figref{fig:interactivetool}).
By placing the radiographs at the detector position and fixing the camera focus point at the position of the x-ray source in the scene, 
the views are consistent with the x-ray projection geometry. 
%The screw is rendered as the silhouette showing the faces perpendicular to the projection direction as lines. 
%Controls allow the selection of the Window and the Level of the radiograph to adapt its contrast.
%The user interacts with the scene by two modes: View Selection to modify the relevant image section and Pose Adaptation to influence the instrument's pose. 
%By dragging the screw users can move the screw to the correct position in parallel to the image plane .
%The rotational degrees of freedom are modified by additional mouse interactions such as right- and middle-click-dragging and the mouse wheel.
Finally, the pose is calculated by tracing multiple points onto the detector (see \cref{sec:radiographs}).

\textbf{i3PosNet: }
Kügler \etal \cite{Kugler.20180226} developed i3PosNet as a general framework to estimate the 
pose of different instruments from single x-rays. 
i3PosNet uses a modular Deep Neural Network and geometric considerations to achieve state-of-the-art pose estimation results from otobasis x-ray images.
In the interest of keeping the paper self-contained, the concept of i3PosNet is briefly introduced.

\textbf{Prediction scheme: }
i3PosNet assumes an initial estimate of the pose is available and only considers the associated part of the image by rotating and cutting an image patch.
This places the expected pose in a well-defined configuration called standard pose.
By predicting the pose on this patch, a better approximation of the pose is found. This pose is used for additional iterations of patchification and pose estimation successively decreasing the difference of the instrument to the standard pose.

The network outputs 6 points in normalized image coordinates.
These points correspond to points defined in the local instrument coordinate system and projected to the image.
Since these points are placed on the main axis and a plane orthogonal to that axis, the position of the instrument's origin is determined by
fitting two lines through said points and calculating their intersection.
The forward angle is derived from the slope of one of the lines.
Although depth and projection angle are resolved by an extension of this method \cite{Kugler.20180226}, they are not part of the analysis in this paper.

\begin{figure}[b]
	\centering
	\subfloat{\includegraphics[width=0.40\textwidth]{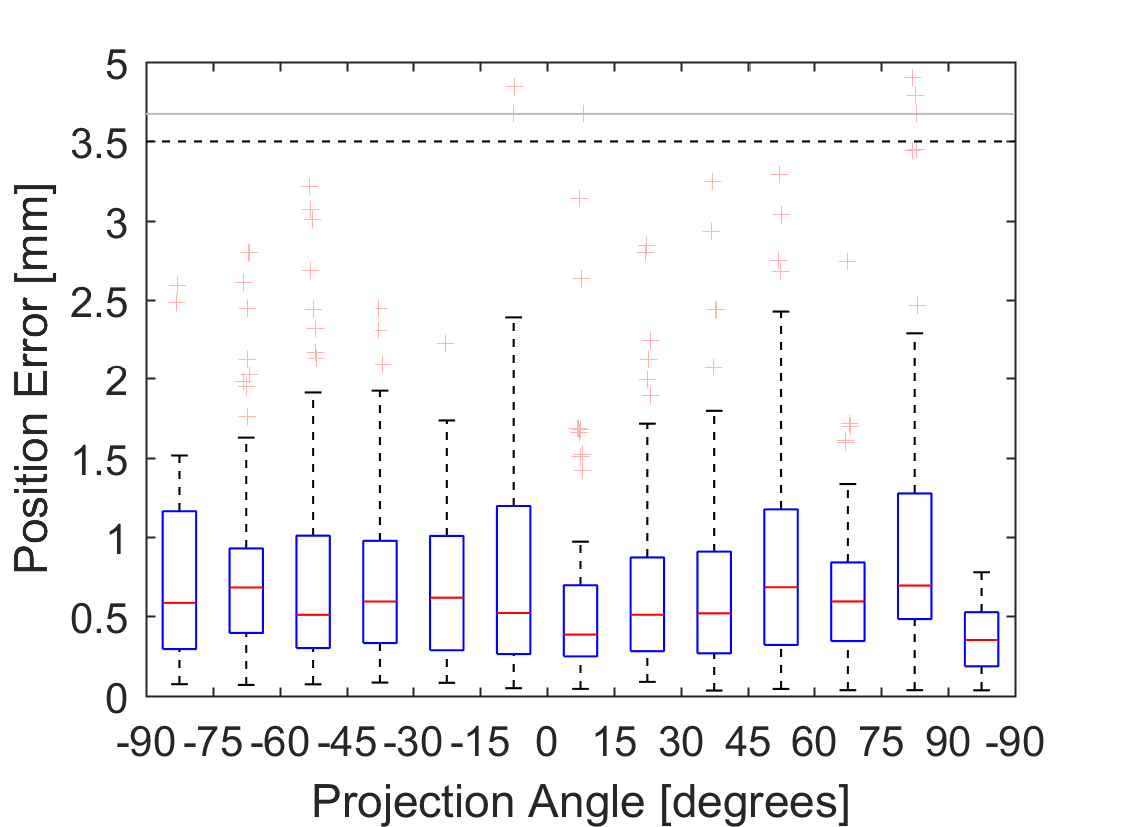}}
	\hspace{5mm}
	\subfloat{\includegraphics[width=0.40\textwidth]{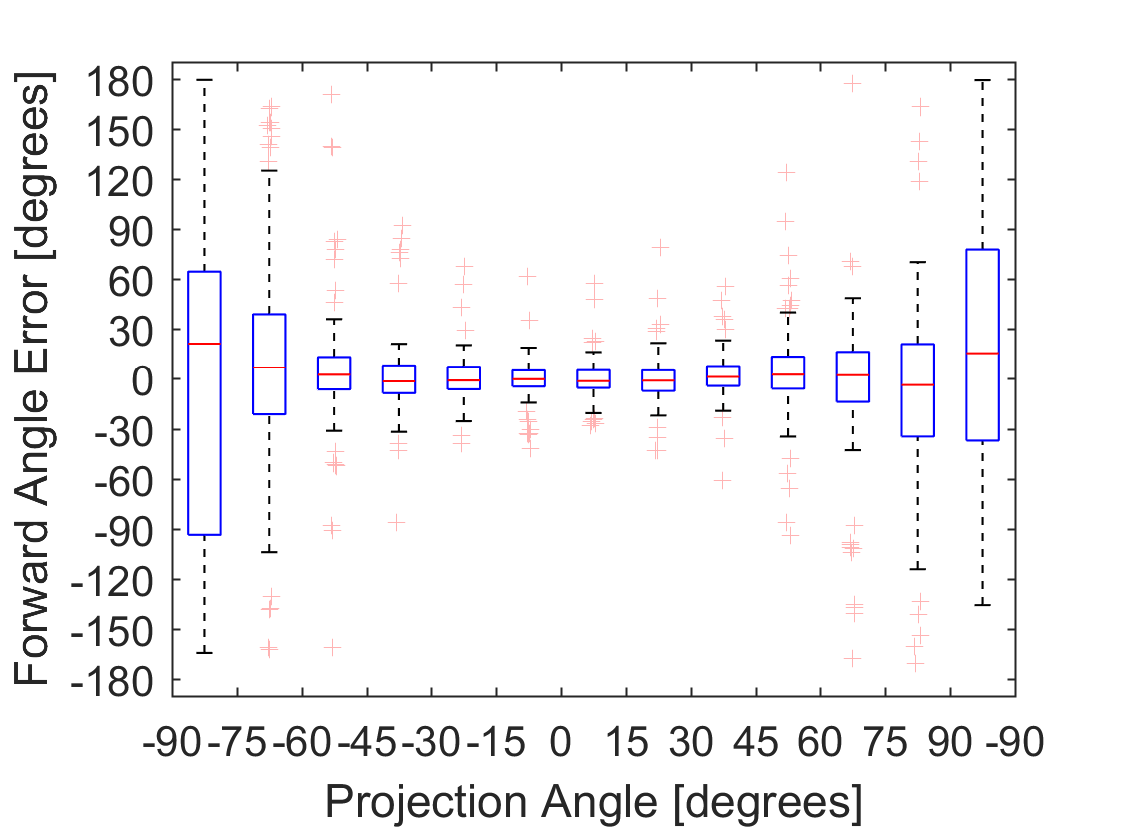}}
	\caption{Evaluation of medical-expert's Position (left) and Forward Angle (right) Error for different projection angles}
	\label{fig:human-performance}       % Give a unique label
\end{figure}

%\begin{figure}[b]
%	\centering
%	\includegraphics[width=0.7\textwidth]{bars/case002.png}
%	\includegraphics[width=0.7\textwidth]{bars/case052.png}
%	\includegraphics[width=0.7\textwidth]{bars/case020.png}
%	\includegraphics[width=0.7\textwidth]{bars/case078.png}
%	\includegraphics[width=0.7\textwidth]{bars/case001.png}
%	\includegraphics[width=0.7\textwidth]{bars/case089.png}
%	\caption{Errors of medical experts for pairs of projections shown at the same time; left: angle errors, right: position errors; color coding: increasing errors from blue to orange, black indicates errors above threshold of \ang{20} for angles and \SI{2}{\milli\meter} for positions}
%	\label{fig:1orientationflawedlabels}       % Give a unique label
%\end{figure}
\textbf{Data Augmentation and Training: }
By adding several random deviations of the pose, multiple augmented patches are derived, which show the instrument with varying offsets. 
In contrast to the superimposed \emph{annotation noise}, these known deviations propagate to the values of the 6 points used as regression targets.
\section{Results}
\label{sec:results}

We analyze the quality of the pose estimation by medical experts as well as the pose estimation of i3PosNet after training on noisy annotations.

\textbf{Error metrics: }
\label{sec:errordefinition}
We evaluate the performance according to two error measures:
\emph{Position Error}: Difference of ground truth and estimated position after projecting them on a plane though the instrument and parallel to the detector measured in Millimeters.
\emph{Forward Angle Error}: Difference of estimated and ground truth angles of the main axis projected onto the image plane.

\textbf{Pose Estimation by medical experts: }
\label{sec:resultsexperts}
We tasked medical experts to annotate our test dataset using the interactive tool for pose estimation presented in \secref{sec:interactivetool}. 
Six young medical professionals (0 to 4 years of medical experience) generated a total of 880 estimates of the poses.

While the estimation of the instrument's position is independent of the projection angle, the instrument's main axis is much harder to identify. 
We observed standard deviations of the individual position components of approximately 4 Pixel leading to average errors of \SI{0.5}{\milli\meter} (see \cref{fig:human-performance}).
Forward Angles showed standard deviations of approximately \ang{15} when looking at the screw from the side to $\ang{70}-\ang{80}$ when being tilted in projection direction.

%\maybedrop{
%	\cref{fig:1orientationflawedlabels} shows typical projections and how they map to errors in terms of forward angles (left) and positions (right).
%}
\begin{figure}[t]
	\centering
	\subfloat{\includegraphics[width=0.4\textwidth]{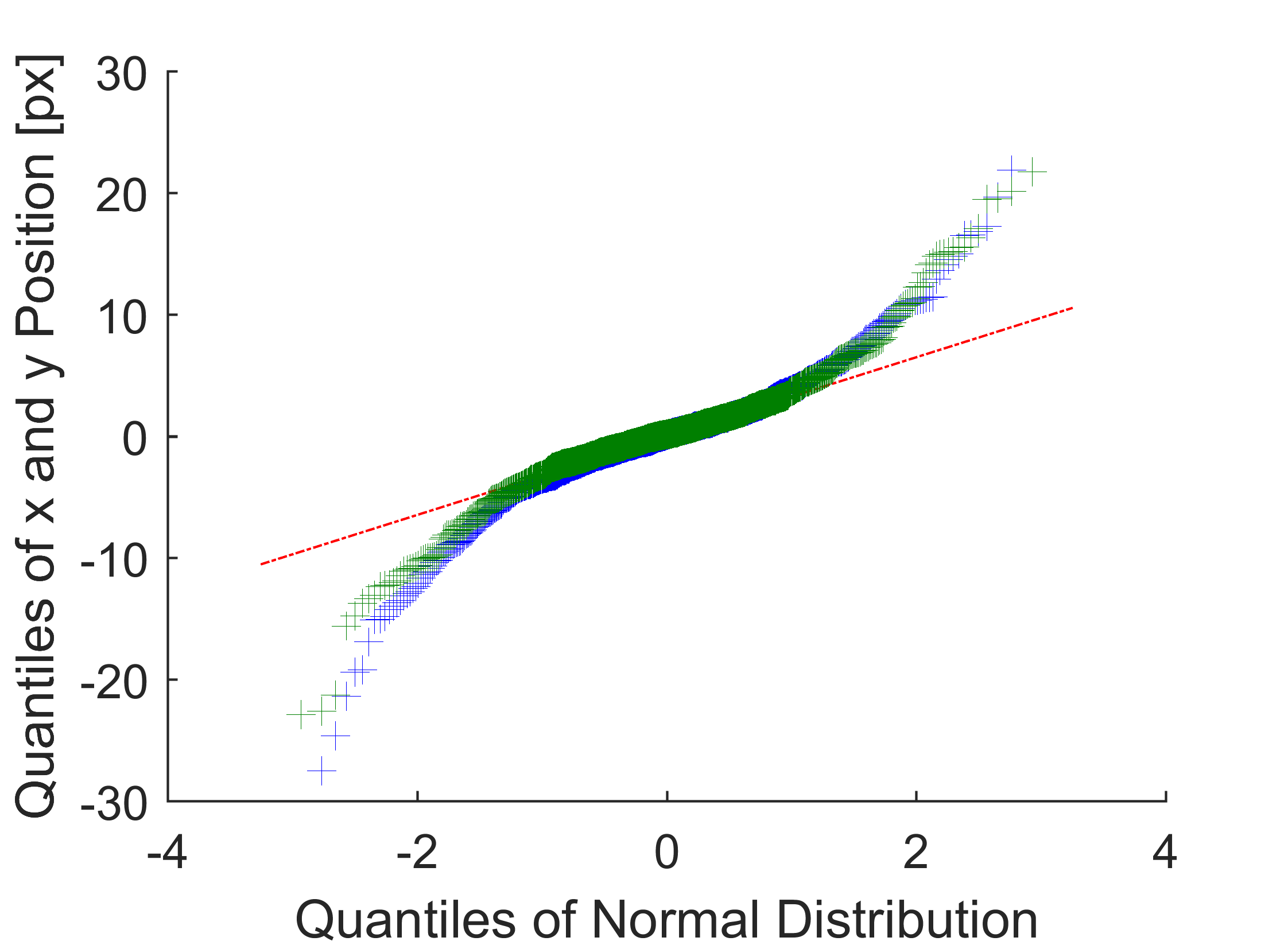}}
	\hspace{5mm}
	\subfloat{\includegraphics[width=0.4\textwidth]{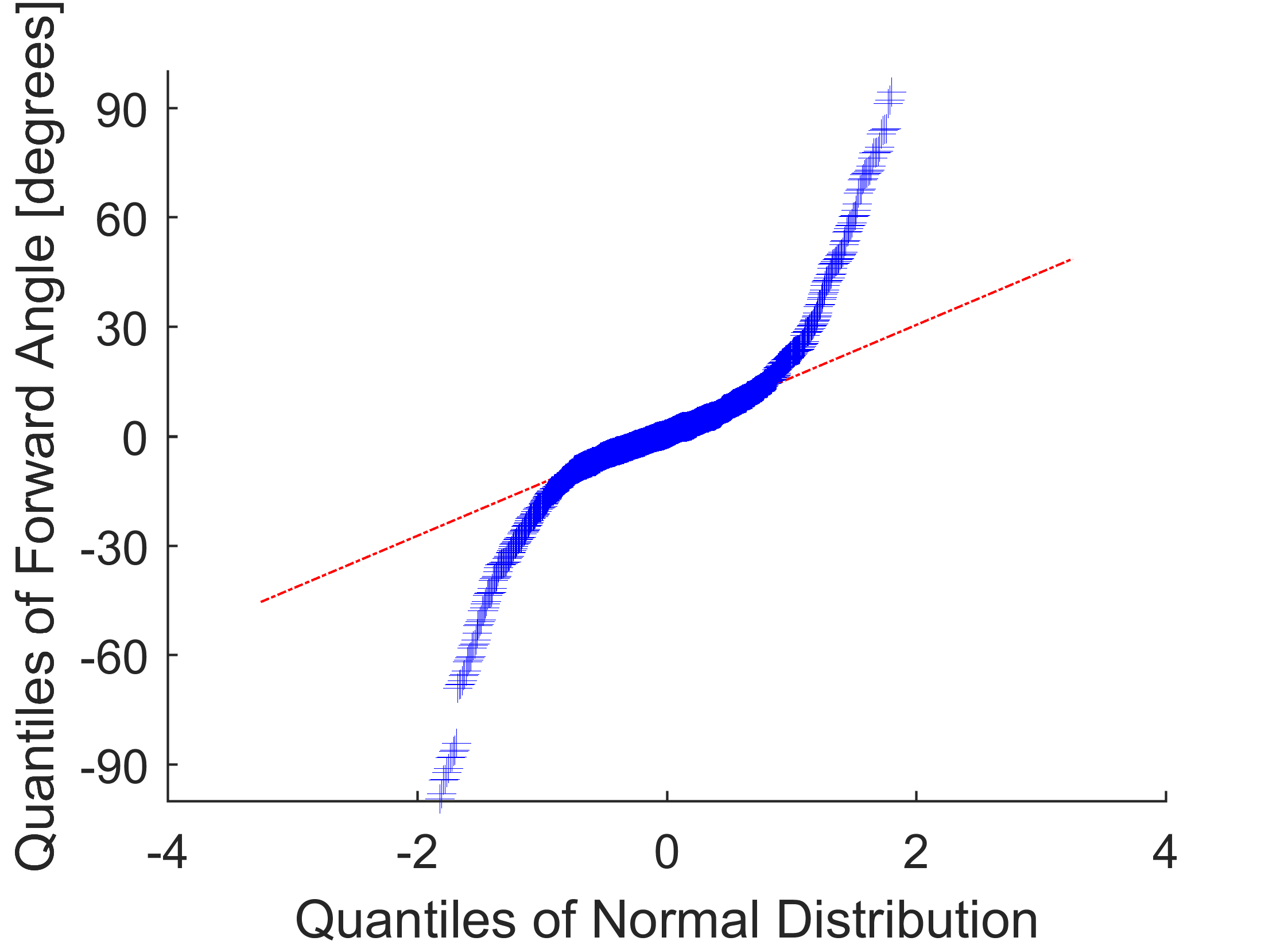}}
	\caption{Comparing the distribution of individual errors with the normal distributionin Q-Q Plots for the Positions (left) and Forward Angle (right) Error}
	\label{fig:qqplots}       % Give a unique label
\end{figure}
To generate larger data sets of ``noisily'' annotated images for training and validating i3PosNet, 
we analyzed the distribution of the errors. 
In Q-Q-Plots, quantiles are mapped to each other for different distributions. 
Accordingly, quantiles mapping to the linear fit (red line) indicates the similarity of the distributions. 
\cref{fig:qqplots} illustrates that a normal distribution represents the experts error distribution very well within 1 to 1.5 standard deviations (70\,\% to 90\,\% of the cases). 
The dataset showed more outliers than expected from a normal distribution, that could be identified manually or by repeated annotation.

\textbf{Impact of the noise level on prediction accuracy: }
We train 5 different models to determine the effect of different noise levels on the prediction quality.
For training, we vary the noise $\eta$ from a minimal configuration (no noise, $\eta=0$) to a maximum ($\eta = 4$), which is similar to annotation noise from manual annotation by medical experts.
Ground truth poses are disturbed by artificial annotation noise drawn from a normal distribution with standard deviations of $(\eta)$ Pixel for the position and $(\eta \cdot{\ang{10}})$ for the forward angle.
Depth and projection angle are perturbed by $(\eta \cdot{\SI{17.5}{\milli\meter}})$ and $(\eta \cdot{\ang{5}})$.
For every image, 20 patches are generated and the model was trained for 40 epochs.
The validation set of 2000 different radiographs was used for \emph{validation} with 2 patches each and identical noise assumptions for annotation.
We found that the model typically started overfitting to the noise after 20 epochs with a trend to overfit earlier for smaller numbers of images.
As a consequence, we chose the model with the lowest validation error for our tests.

\begin{figure}[h]
	\centering
	\includegraphics[width=0.5\textwidth]{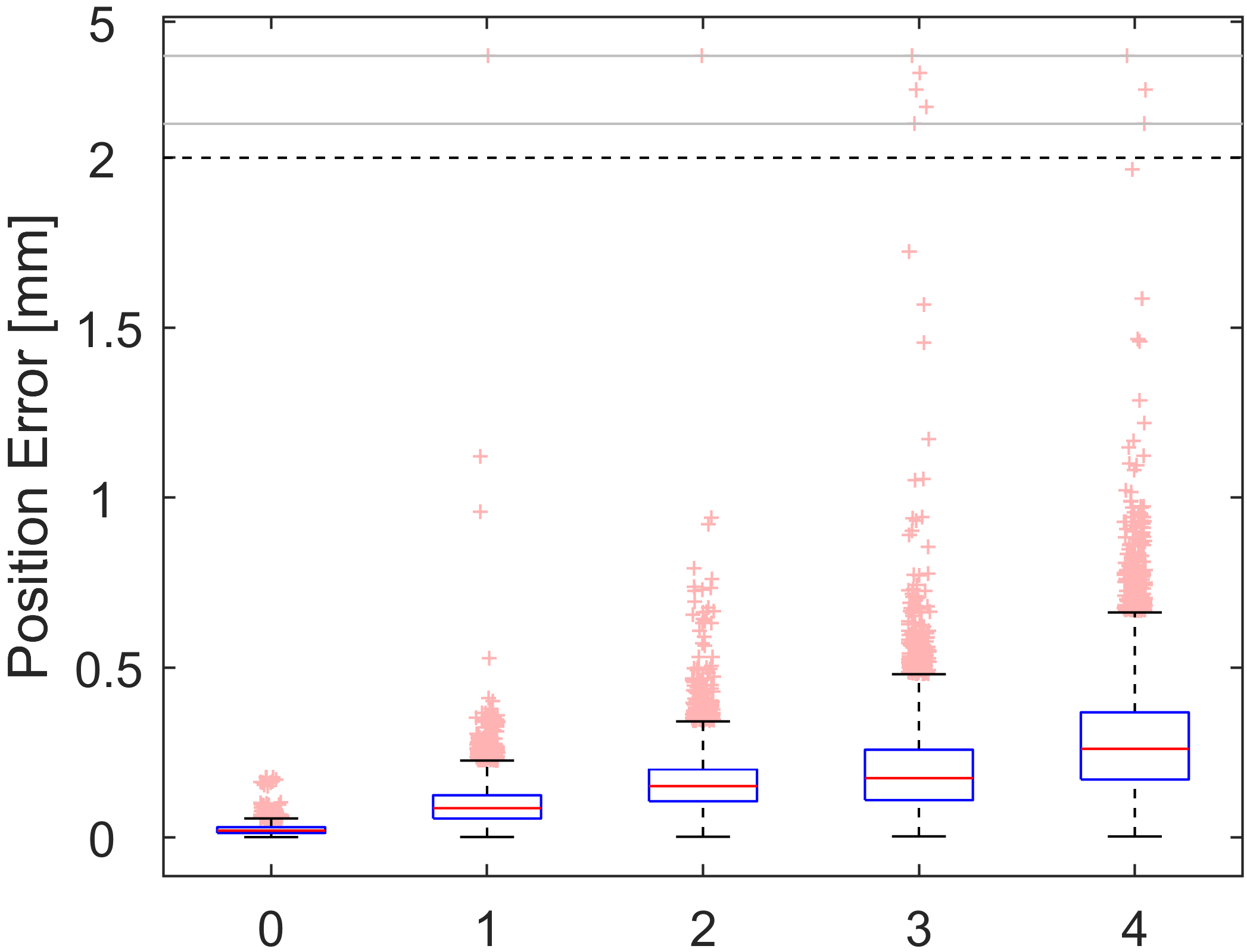}\includegraphics[width=0.5\textwidth]{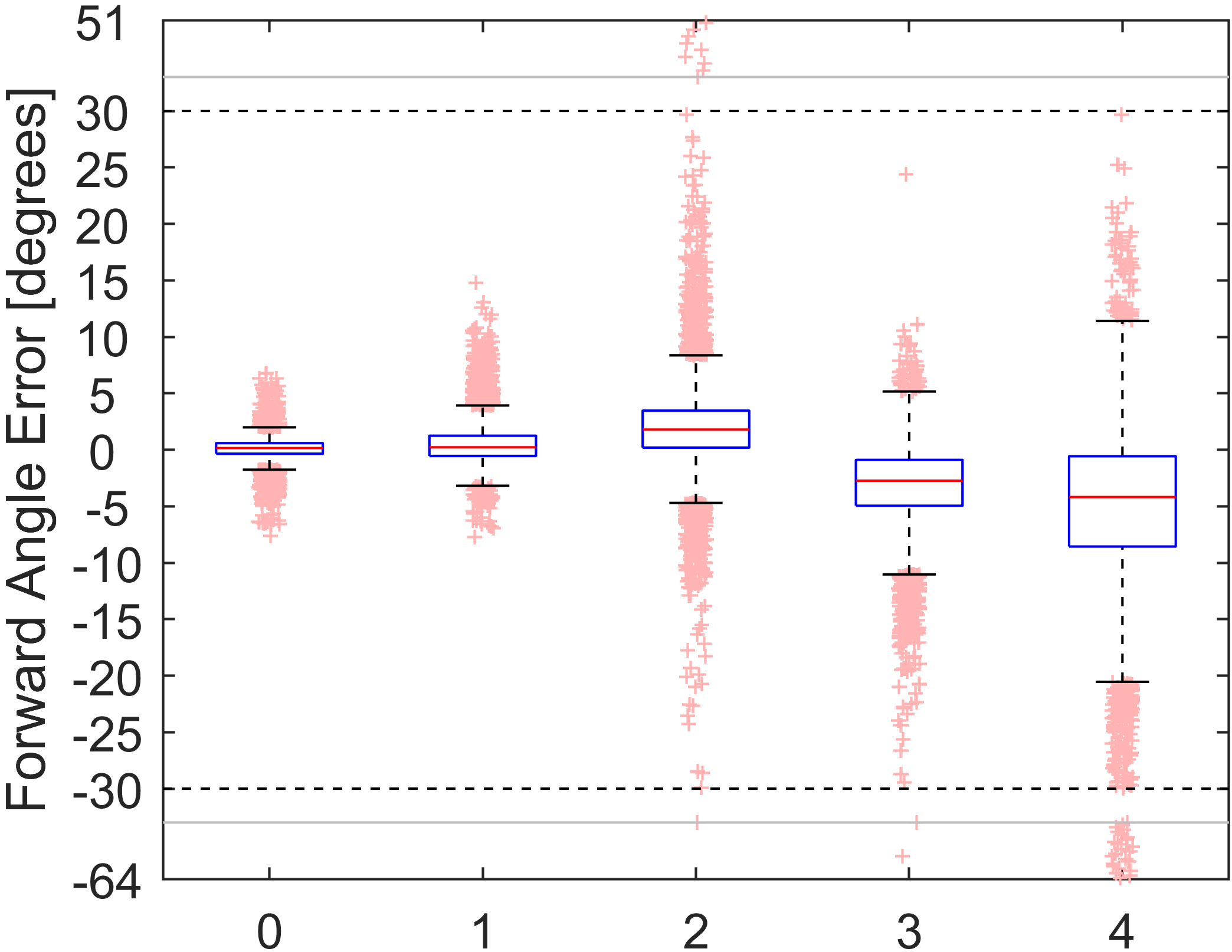}
	\caption{\label{fig:i3posnet-noise}Comparisions of Position (left) and Forward Angle (right) Errors of i3PosNet for different noise levels; Position: lower is better,  Forward Angle: smaller boxes centered around zero are better}       % Give a unique label
\end{figure}
All models are tested on 1000 unseen images (from the unseen anatomy) with 10 repetitions each.
The results (see \cref{fig:i3posnet-noise}) indicate a linear relationship between the noise level and the prediction accuracy.

\begin{figure}[t]
	\centering
	\includegraphics[width=0.5\textwidth]{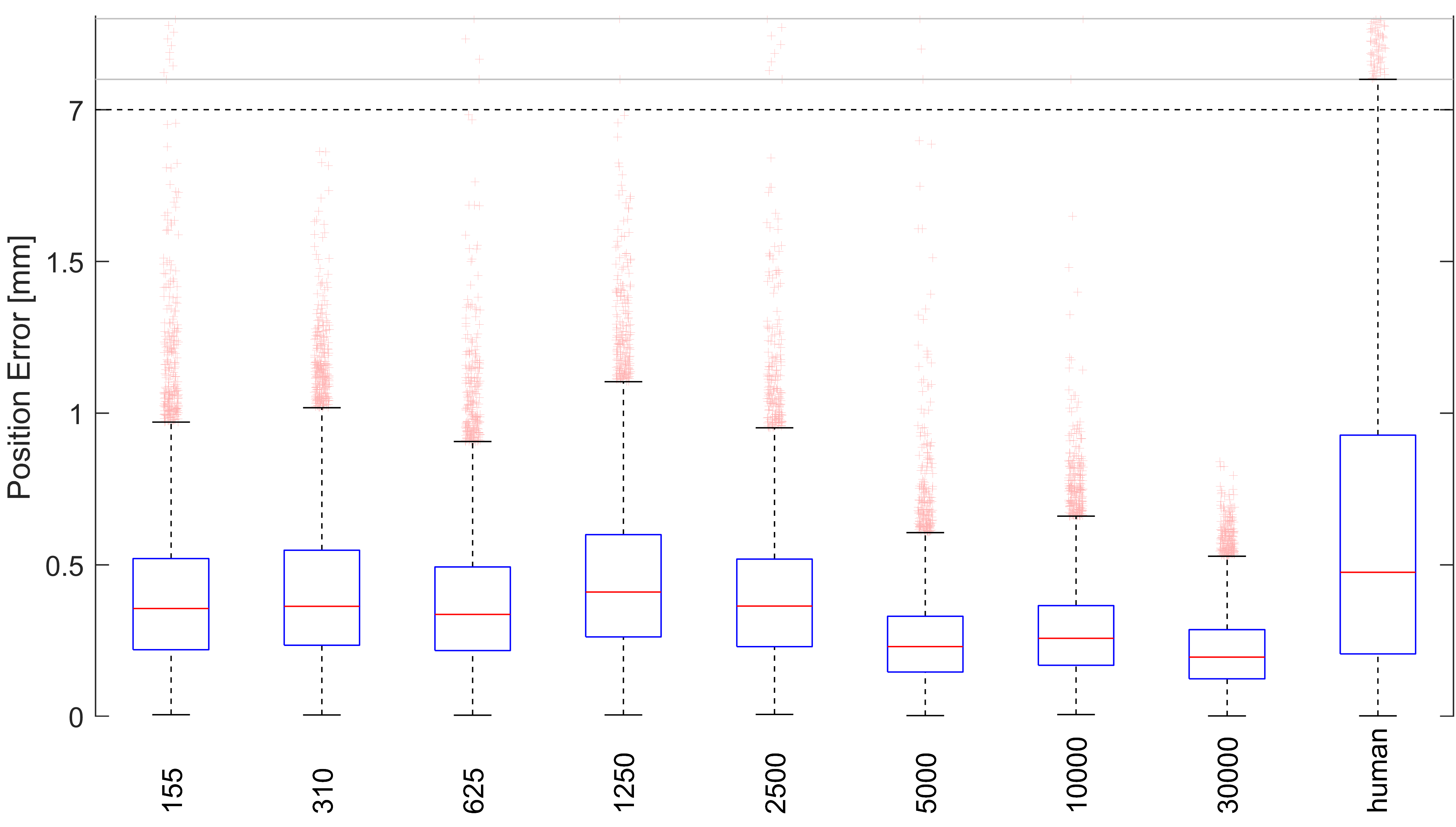}\includegraphics[width=0.5\textwidth]{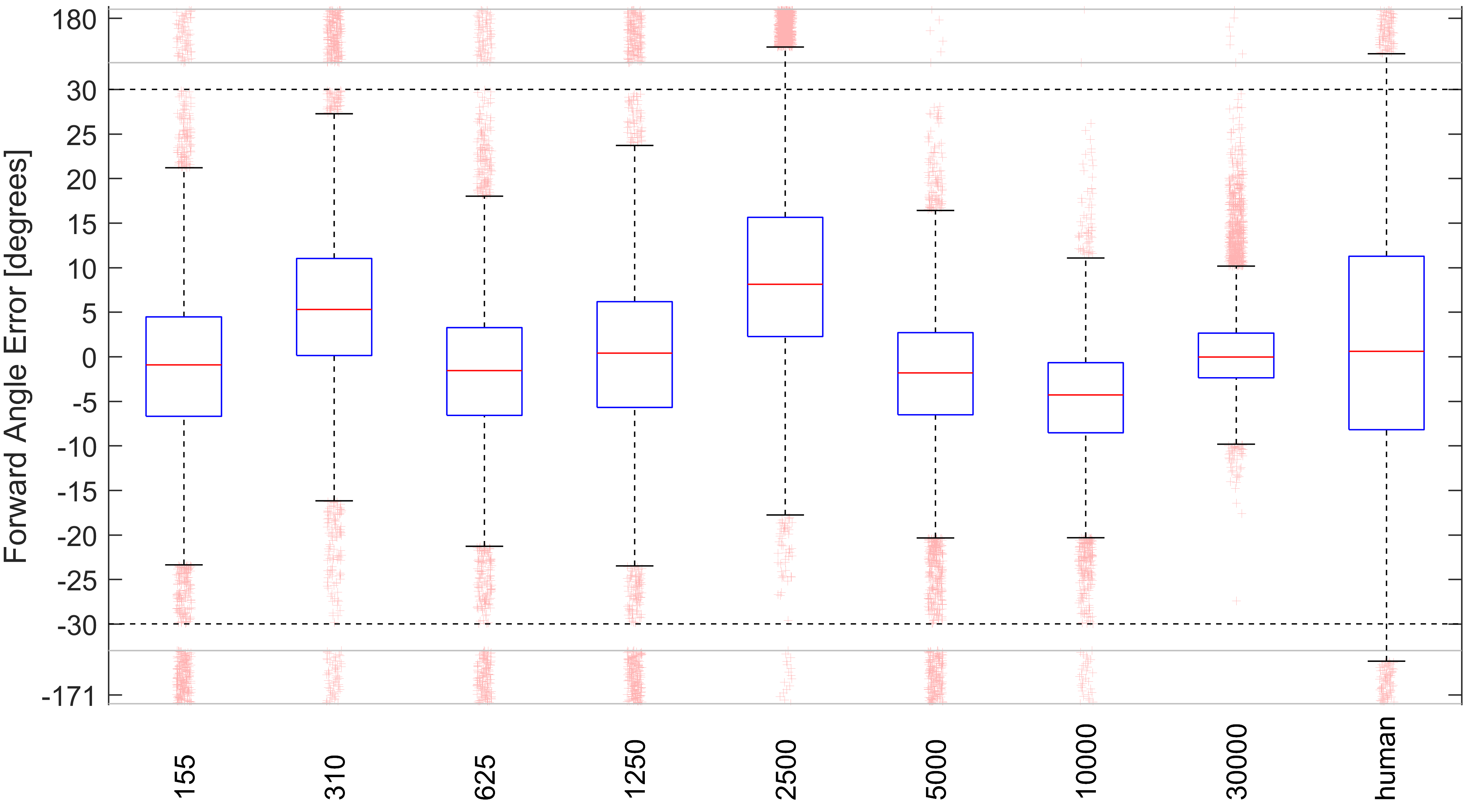}
	\caption{\label{fig:human-vs-i3posnet}Comparing Position (left) and Forward Angle (right) Errors for i3PosNet and medical experts; Position: lower is better,  Forward Angle: smaller boxes centered around zero are better}       % Give a unique label
\end{figure}

\textbf{Comparison of medical experts and i3PosNet: }
To identify how much noisy annotation is enough to outperform medical experts,
we evaluate 8 scenarios for i3PosNet with different numbers of unique images used for training: 155, 310, 625, 1250, 5000, 10000 and the triple-annotation of the 10000 annotations (represented as 30000 in \cref{fig:human-vs-i3posnet}).
All scenarios feature the noise level derived from the manual annotation experiment (noise 4:, 4 Pixel) and follow the same generation procedure.
The number of training epochs is normalized to the 10000 image case, i.e. all scenarios underwent the same number of gradient updates.
%\begin{figure}[t]
%	\centering
%	\caption{\label{fig:human-vs-i3posnet-mm}Comparison of the  of i3PosNet trained on different numbers of noisy annotations and medical experts (lower is better)}       % Give a unique label
%\end{figure}

For every scenario, 1000 images are tested with 10 repetitions.
Test errors decreased as the number of images increased and exceeded the training noise level at 5000 images for position and forward angle (\cref{fig:human-vs-i3posnet}) errors.

\section{Discussion}
\label{sec:discussion}
	We systematically evaluated the instrument pose annotation ability of medical experts and found errors at a level larger than anticipated.
	Training i3PosNet with statistically generated noisy annotations, we found a linear relationship between the noise level and the prediction accuracy. 
	For noise derived from human annotations, we found that for 5000 images and more i3PosNet consitently generates better predictions than the input noise and outperformed medical experts, i.e. i3PosNet developed a generalized representation over annotations.
	
	Despite the increase in performance over medical experts, 
	in many cases MIS applications such as cochlear implant surgery require significantly better error levels.
	We plan to improve i3PosNet introducing Semi- and Unsupervised Learning schemes to become independent of external annotations.
	
%	\needswork{
	The robustness of Deep Learning to noisy annotations documented here opens up new research opportunities for systematic regression studies in MIA.
%	}

%While the former inherently carry ground truth uncertainties leading to inaccurate ground truth poses, the generalization of methods from artificial to real images has to be proven by additional experiments.	

	\renewcommand{\bibsection}{\section*{References}} % requried for natbib to have "References" printed and as section*, not chapter*
	% Use natbib compatbile splncsnat style.
	% It does provide all features of splncs03, but is developed in a clean way.
	% Source: http://phaseportrait.blogspot.de/2011/02/natbib-compatible-bibtex-style-bst-file.html
	\bibliographystyle{splncsnat}
	\begingroup
	\microtypecontext{expansion=sloppy}
	\bibliography{bibliography}
	\endgroup
\end{document}